\documentclass[conference]{IEEEtran}
\IEEEoverridecommandlockouts
\usepackage{cite}
\usepackage{amsmath,amssymb,amsfonts}
\usepackage{algorithmic}
\usepackage{graphicx}
\usepackage{textcomp}
\usepackage{xcolor}
\def\BibTeX{{\rm B\kern-.05em{\sc i\kern-.025em b}\kern-.08em
    T\kern-.1667em\lower.7ex\hbox{E}\kern-.125emX}}
\begin{document}

\title{Comparative Study of Multi-Agent Actor-Critic Algorithms in Parameterized Action Reinforcement Learning\\
}

\author{\IEEEauthorblockN{Ubayd Ali Bapoo}
\IEEEauthorblockA{\textit{Department of Computer Science} \\
\textit{University of the Western Cape}\\
Cape Town, South Africa \\
Email: 2934126@myuwc.ac.za}
\and
\IEEEauthorblockN{Clement N Nyirenda}
\IEEEauthorblockA{\textit{Department of Computer Science} \\
\textit{University of the Western Cape}\\
Cape Town, South Africa \\
ORCID: 0000-0002-4181-0478}
}

\IEEEpubid{\makebox[\columnwidth]{979-8-3195-0703-7/26/\$31.00~\copyright2026 IEEE \hfill}
\hspace{\columnsep}\makebox[\columnwidth]{}}

\maketitle

\IEEEpubidadjcol

\begin{abstract}

Parameterized action reinforcement learning has shown strong performance in environments requiring both discrete action selection and continuous parameterization. Prior work established the effectiveness of single-agent actor-critic algorithms — Greedy Actor-Critic (GAC), Soft Actor-Critic (SAC), and Truncated Quantile Critics (TQC) — on benchmark parameterized action tasks, but their extension to multi-agent settings remains largely unexplored. This paper presents a comparative study of shared-experience multi-agent extensions of these algorithms: Multi-Agent Greedy Actor-Critic (MAGAC), Multi-Agent Soft Actor-Critic (MASAC), and Multi-Agent Truncated Quantile Critics (MATQC). Rather than following the centralized training, decentralized execution (CTDE) paradigm, the proposed framework uses multiple independent actor-critic agents that share a replay buffer while maintaining separate policy and value networks. We evaluate the algorithms on the Platform-v0 and Goal-v0 benchmarks against their single-agent counterparts, using three-, five-, and ten-agent configurations to assess scalability. Performance is measured by average evaluation return and training time across ten independent runs, with one-way ANOVA and Tukey HSD post-hoc tests used to assess statistical significance. Results show that the multi-agent framework consistently improves Greedy Actor-Critic performance, while MASAC and MATQC show comparatively modest gains over their single-agent versions. Increasing the number of agents beyond five yields limited additional performance while substantially raising computational cost, particularly for MAGAC. These results highlight a trade-off between learning performance and computational efficiency, offering insight into the scalability of shared-experience multi-agent actor-critic methods for parameterized action reinforcement learning.

\end{abstract}

\begin{IEEEkeywords}
Parameterized action reinforcement learning, multi-agent reinforcement learning, actor-critic methods, soft actor-critic, hybrid action spaces, shared-experience multi-agent learning
\end{IEEEkeywords}

\section{Introduction}
Search engines have become a fundamental component of modern information retrieval systems, enabling users to efficiently access information from vast collections of digital documents. Over the past two decades, significant advancements have been made in the architecture and performance of large-scale search engines, particularly in the areas of indexing techniques, query processing, and ranking mechanisms. One of the most critical technologies supporting efficient information retrieval is the inverted index, which allows search engines to rapidly identify documents containing specific query terms while maintaining scalability across billions of web pages \cite{zobel2006}. Despite these advancements, search engines still face challenges in balancing the competing objectives of retrieval quality and response latency when processing user queries.

As the volume of information on the web continues to grow exponentially, processing all possible candidate documents for every query becomes computationally expensive. Modern search engines therefore employ sophisticated mechanisms to reduce the number of candidate documents that must be considered during ranking. One such mechanism is the use of match plans, which define a sequence of matching rules and stopping conditions that guide the process of document retrieval from the inverted index. Match plans determine how posting lists from different document fields are accessed and combined during query evaluation, allowing search engines to efficiently generate candidate documents while minimizing computational overhead \cite{rosset2018optimizing}. These match rules essentially control how deeply the search engine scans the index and when the retrieval process should terminate.

Traditionally, match plans have been designed manually by engineers based on domain expertise and empirical experimentation. Engineers typically analyze system performance and construct match rules that balance the trade-off between search accuracy and system efficiency. However, manually designed match plans can be difficult to optimize, particularly in large-scale search environments where query patterns and document distributions change dynamically. This limitation has motivated research into automated methods for generating and optimizing match plans using machine learning and reinforcement learning techniques \cite{luo2021matchplan}.

Reinforcement learning (RL) has emerged as a powerful framework for solving sequential decision-making problems where an agent learns optimal behaviour through interactions with an environment. In an RL setting, the agent observes the current state of the environment, selects an action, and receives feedback in the form of a reward signal that reflects the quality of the action taken. Through repeated interactions with the environment, the agent learns policies that maximize cumulative rewards over time, following the standard reinforcement learning framework described by \cite{ladosz2022exploration}. Reinforcement learning is particularly well suited to problems involving sequential decision processes, such as query evaluation strategies in search engines, where decisions must be made step-by-step to optimize overall system performance.

Among the various reinforcement learning approaches, actor–critic algorithms have gained significant attention due to their ability to combine the advantages of policy-based and value-based learning methods. In the actor–critic framework, the actor is responsible for selecting actions based on a learned policy, while the critic evaluates the quality of those actions using a value function. This interaction between the actor and the critic allows the learning process to be more stable and efficient compared to purely policy-based or value-based approaches \cite{chen2020songbird, mnih2016a3c}. Actor–critic methods have therefore become widely used in complex reinforcement learning tasks that involve high-dimensional state and action spaces.

Recent research has proposed several actor–critic algorithms that improve learning stability, exploration capability, and overall performance in reinforcement learning environments. One such algorithm is the Soft Actor-Critic (SAC), which introduces an entropy maximization objective into the reinforcement learning framework. By encouraging exploration through entropy regularization, SAC is able to learn more robust policies and avoid premature convergence to suboptimal solutions \cite{ding2021averaged}. In addition to SAC, other actor–critic variants have been proposed to address specific challenges in reinforcement learning. For example, the Greedy Actor-Critic (GAC) algorithm improves policy optimization by selecting high-value actions through a cross-entropy method \cite{neumann2023greedy}, while the Truncated Quantile Critics (TQC) algorithm mitigates overestimation bias by employing distributional reinforcement learning techniques \cite{foo2023drl}.

Many real-world decision-making problems involve environments where actions consist of both discrete choices and continuous parameters. These environments are known as parameterized action spaces and pose additional challenges for reinforcement learning algorithms due to their hybrid action structure. Parameterized reinforcement learning methods extend traditional RL frameworks to handle such environments by allowing agents to select discrete actions together with associated continuous parameters. Benchmark environments such as the Platform and Goal domains have been widely used to evaluate the performance of reinforcement learning algorithms in parameterized action spaces \cite{bester2019multipass}.

The work in \cite{bapoo2025icit} investigated the application of parameterized action actor–critic reinforcement learning algorithms for match plan generation in web search systems. In particular, Parameterized Action Soft Actor-Critic (PASAC), Parameterized Action Greedy Actor-Critic (PAGAC), and Parameterized Action Truncated Quantile Critics (PATQC) were evaluated in benchmark environments to analyze their performance in complex decision-making tasks. The results demonstrated that alternative actor–critic algorithms such as GAC and TQC can outperform SAC in certain scenarios, particularly in terms of training efficiency and evaluation performance. These findings highlight the importance of exploring different reinforcement learning approaches when optimizing sequential decision processes related to search engine efficiency.

While these studies provide valuable insights into the behaviour and performance of single-agent reinforcement learning algorithms, they do not consider the potential advantages of multi-agent reinforcement learning (MARL). In many complex systems, multiple agents can collaborate to solve tasks more effectively by sharing information and coordinating their actions. Multi-agent reinforcement learning enables multiple decision-making entities to interact within a shared environment, allowing them to learn cooperative strategies that improve overall system performance.

Recent advances in multi-agent reinforcement learning have demonstrated that extending actor–critic algorithms to multi-agent settings can significantly improve system robustness and scalability. One such approach is the Multi-Agent Soft Actor-Critic (MASAC) algorithm, which extends SAC within a centralized training and decentralized execution framework. In this architecture, agents are trained collaboratively using shared information during the learning phase while executing policies independently during deployment. This structure allows agents to coordinate their behaviour while maintaining scalability in complex environments \cite{xie2024masac}. Studies have shown that MASAC can improve stability and performance in multi-agent control systems, particularly in environments that require cooperative decision-making among multiple agents.

Motivated by these developments, this study extends prior research \cite{bapoo2025icit} by examining the application of multi-agent actor–critic algorithms. Specifically, it investigates the performance of multi-agent variants of SAC, GAC, and TQC within the same benchmark environments used in previous single-agent studies \cite{bapoo2025icit,bester2019multipass}. By introducing cooperative learning among multiple agents, the proposed approach aims to improve coordination, scalability, and performance in parameterized action reinforcement learning environments. The results of this study provide insights into the effectiveness of multi-agent reinforcement learning for complex decision-making problems and contribute to the development of more efficient reinforcement learning strategies for large-scale systems. Furthermore, the scalability of the proposed framework is investigated by evaluating configurations consisting of three, five and ten agents.

\subsection{Contributions}

This paper presents an empirical extension of parameterised action actor--critic reinforcement learning algorithms to a multi-agent setting.

Unlike existing CTDE-based approaches, including MADDPG \cite{lowe2017multiagent}, QMIX \cite{rashid2018qmix}, and MAPPO \cite{yu2022mappo}, the proposed framework employs independent actor--critic agents that learn through a shared replay buffer while maintaining separate policy and value networks.

The contribution is therefore not the development of a new theoretical MARL algorithm, but rather a controlled empirical investigation of how shared-experience multi-agent learning influences actor--critic methods originally designed for parameterised action spaces. The main contributions of this work are as follows:
\begin{itemize}
    \item The extension of PASAC, PAGAC, and PATQC to shared-experience multi-agent variants, namely MASAC, MAGAC, and MATQC.
    \item A controlled comparison between single-agent and multi-agent formulations using the same benchmark environments, reward structure, and training configuration.
    \item An agent-scaling analysis that evaluates the effect of increasing the number of agents, using 3-agent, 5-agent, and 10-agent configurations.
    \item A statistical evaluation of algorithm performance using one-way ANOVA and Tukey HSD post-hoc testing.
    \item An analysis of both evaluation return and training time to assess the trade-off between learning performance and computational efficiency.
\end{itemize}

Although this work is motivated by centralized training and decentralized execution (CTDE) approaches in MARL, the implemented framework should be interpreted as a shared-experience multi-agent extension rather than a full CTDE architecture. In the implemented framework, multiple actor--critic agents are instantiated and trained using shared experience replay, while action selection is performed through action fusion. A full CTDE formulation with a centralized critic over joint observations and joint actions is left for future work.

The remainder of this paper is organised as follows. Section II presents the background and related work. Section III describes the simulation setup. Section IV presents the results. Sections V and VI present the discussion and conclusion, respectively.

\section{Background and Related Work}

This section provides an overview of the research areas most relevant to this work, including match plan generation in web search systems, reinforcement learning methods for sequential decision-making, parameterised action reinforcement learning, and multi-agent reinforcement learning approaches.

The work in \cite{bapoo2025icit} investigated the performance of parameterised action reinforcement learning algorithms, including Parameterised Soft Actor-Critic (PASAC), Parameterised Greedy Actor-critic (PAGAC) and Parameterised Truncated Quantitative Critics (PATQC). The study evaluated these algorithms across the Platform and Goal benchmark environments and analysed their performance in terms of evaluation return and training efficiency. The results indicated that Greedy Actor-Critic-based approaches achieved strong performance compared to the other evaluated algorithms. Building on this work, the present study extends these algorithms to a multi-agent framework in order to investigate their behaviour and performance in multi-agent reinforcement learning environments.

\subsection{Search Engine Retrieval and Match Plan Generation}

Modern web search engines must process queries across extremely large collections of documents while maintaining high retrieval quality and low response times. Efficient document retrieval is typically achieved through inverted indexes, which allow search engines to quickly locate documents containing specific query terms. Inverted indexing techniques have been widely studied and remain a fundamental component of large-scale search systems \cite{zobel2006}.

During query evaluation, search engines must generate a set of candidate documents that will later be ranked by machine learning models. However, processing all documents associated with query terms would be computationally infeasible. To address this challenge, search engines employ mechanisms such as match plans to guide the retrieval process. Match plans define a sequence of rules that determine how posting lists are accessed, how documents are matched against query terms, and when the retrieval process should terminate \cite{rosset2018optimizing}. These rules enable search engines to balance the trade-off between retrieval effectiveness and system efficiency.

Traditionally, match plans have been manually designed by engineers based on their knowledge of the search system and empirical experimentation. However, manually constructed match plans are difficult to optimize and may not adapt well to changing query patterns and document distributions. To overcome these limitations, recent research has explored the use of reinforcement learning techniques to automatically generate match plans. Luo et al. proposed a reinforcement learning framework for match plan generation that models the retrieval process as a sequential decision-making problem, allowing an agent to learn optimal match rules based on observed system performance \cite{luo2021matchplan}.

\subsection{Reinforcement Learning and Actor–Critic Algorithms}

Reinforcement learning has become an important paradigm for solving complex decision-making problems where agents must learn optimal actions through interactions with an environment. In this framework, the agent observes the state of the environment, performs an action, and receives a reward signal that indicates the quality of the action taken. Through repeated interactions, the agent learns policies that maximize cumulative rewards over time \cite{ladosz2022exploration}. Among reinforcement learning methods, actor–critic algorithms have gained significant attention due to their ability to combine policy-based learning with value-based evaluation. In this framework, the actor determines the policy used to select actions, while the critic evaluates the expected return of those actions. This interaction allows the learning process to be more stable and efficient compared to traditional reinforcement learning approaches \cite{chen2020songbird}.

Several actor–critic algorithms have been proposed to improve learning performance in complex environments. The Soft Actor-Critic (SAC) algorithm introduces entropy regularization into the reinforcement learning objective to encourage exploration and improve policy robustness \cite{ding2021averaged}. Other approaches such as the Greedy Actor-Critic (GAC) algorithm improve policy optimization by selecting high-value actions through a cross-entropy based method \cite{neumann2023greedy}. Additionally, the Truncated Quantile Critics (TQC) algorithm employs distributional reinforcement learning techniques to reduce overestimation bias in value estimation and improve policy stability \cite{foo2023drl}.

\subsection{Parameterized Action Reinforcement Learning}

Many real-world reinforcement learning problems involve hybrid action spaces consisting of both discrete actions and continuous parameters. These environments are referred to as parameterised action spaces and present additional challenges for reinforcement learning algorithms due to the complexity of selecting both the action type and its associated parameters. To address this challenge, parameterised action reinforcement learning methods have been proposed to extend traditional reinforcement learning algorithms to hybrid action spaces. Benchmark environments such as the Platform and Goal domains have been widely used to evaluate reinforcement learning algorithms operating in parameterised action spaces. These environments require agents to choose discrete actions such as movement types while simultaneously selecting continuous parameters controlling the magnitude or direction of the action \cite{bester2019multipass, xiong2018pdqn}.

The study in \cite{bapoo2025icit} previously examined parameterised action actor–critic algorithms for match plan generation tasks. In particular, PASAC, PAGAC, and PATQC were evaluated in benchmark environments with parameterised action spaces. The results showed that these methods can successfully learn effective policies in hybrid action settings and enhance decision-making efficiency in complex sequential tasks.

\subsection{Multi-Agent Reinforcement Learning}
While much reinforcement learning research focuses on single-agent environments, many real-world problems involve multiple agents interacting within a shared environment. Multi-agent reinforcement learning (MARL) extends traditional reinforcement learning frameworks to scenarios in which multiple agents learn simultaneously, either cooperating or competing with one another. Such settings arise naturally in domains including autonomous vehicle coordination, smart grid management, and robotic swarm control, where the behaviour of each agent is inherently coupled with the actions of others. A central challenge in MARL is environmental non-stationarity. Because multiple agents learn concurrently, changes in one agent's policy alter the environment perceived by others, complicating stable learning and hindering convergence. This issue is further compounded in environments with sparse rewards or delayed feedback, where distinguishing the contribution of individual agents to a shared outcome becomes particularly difficult. This problem, commonly referred to as the credit assignment problem, remains an active area of research in the multi-agent learning community.

A prominent approach to addressing non-stationarity is the centralized training with decentralized execution (CTDE) framework, in which agents share information during training but operate independently at deployment. Representative CTDE algorithms include MADDPG, QMIX and MAPPO, which have demonstrated strong performance across a variety of cooperative multi-agent benchmark tasks \cite{rashid2018qmix, yu2022mappo}. This design enables agents to learn cooperative strategies while preserving scalability in large systems. During centralised training, agents can access the global state and the policies of other agents, providing richer learning signals and reducing the effective non-stationarity of the environment. At execution time, however, each agent relies only on its local observations, ensuring that the framework remains practical in settings where full state information is unavailable or communication between agents is restricted.

Building directly on the CTDE paradigm, the Multi-Agent Soft Actor-Critic (MASAC) algorithm extends the SAC framework to multi-agent settings and has demonstrated improved stability and robustness in cooperative control tasks \cite{xie2024masac}. MASAC retains the maximum entropy objective of SAC, encouraging agents to maintain diverse and exploratory policies, which is particularly beneficial in cooperative settings where premature convergence to suboptimal joint policies is a common failure mode. Notably, Xie et al. \cite{xie2024masac} applied MASAC to multi-microgrid systems, where multiple agents coordinate to maintain frequency stability across interconnected microgrids. Their results show that coordinated decision-making under the MASAC framework meaningfully improves system robustness and disturbance rejection.

Despite these advances, relatively little work has explored multi-agent actor–critic methods in parameterised action spaces, where agents must jointly select both discrete action types and associated continuous parameters. This added complexity introduces additional coordination challenges, as agents must align not only their high-level action choices but also the continuous parameters that govern their execution. This research therefore investigates whether extending algorithms such as SAC, GAC, and TQC to a multi-agent framework can improve learning performance and coordination in parameterised action reinforcement learning environments.

\section{Proposed Multi-Agent Framework}
The proposed framework extends the parameterised action actor--critic algorithms investigated in previous work to a shared-experience multi-agent architecture. Rather than employing a single learning agent, multiple independent actor--critic agents are instantiated within the same parameterised action environment. Each agent maintains its own actor and critic networks while interacting with a common environment during training.

During training, the agents generate experience through repeated interaction with the environment. The generated transitions are stored in a shared replay buffer, allowing each agent to learn from experiences collected across the group. This shared replay mechanism increases the diversity of training samples while preserving independent policy and value-function representations for each agent. Each agent independently selects a parameterised action consisting of a discrete action and its associated continuous parameters. The individual actions are then combined using an action fusion strategy to produce a single executable action for the environment. This enables multiple agents to contribute to the decision-making process while remaining compatible with benchmark environments that accept one parameterised action at each timestep.

\begin{figure}[ht]
\centering
\includegraphics[width=\linewidth]{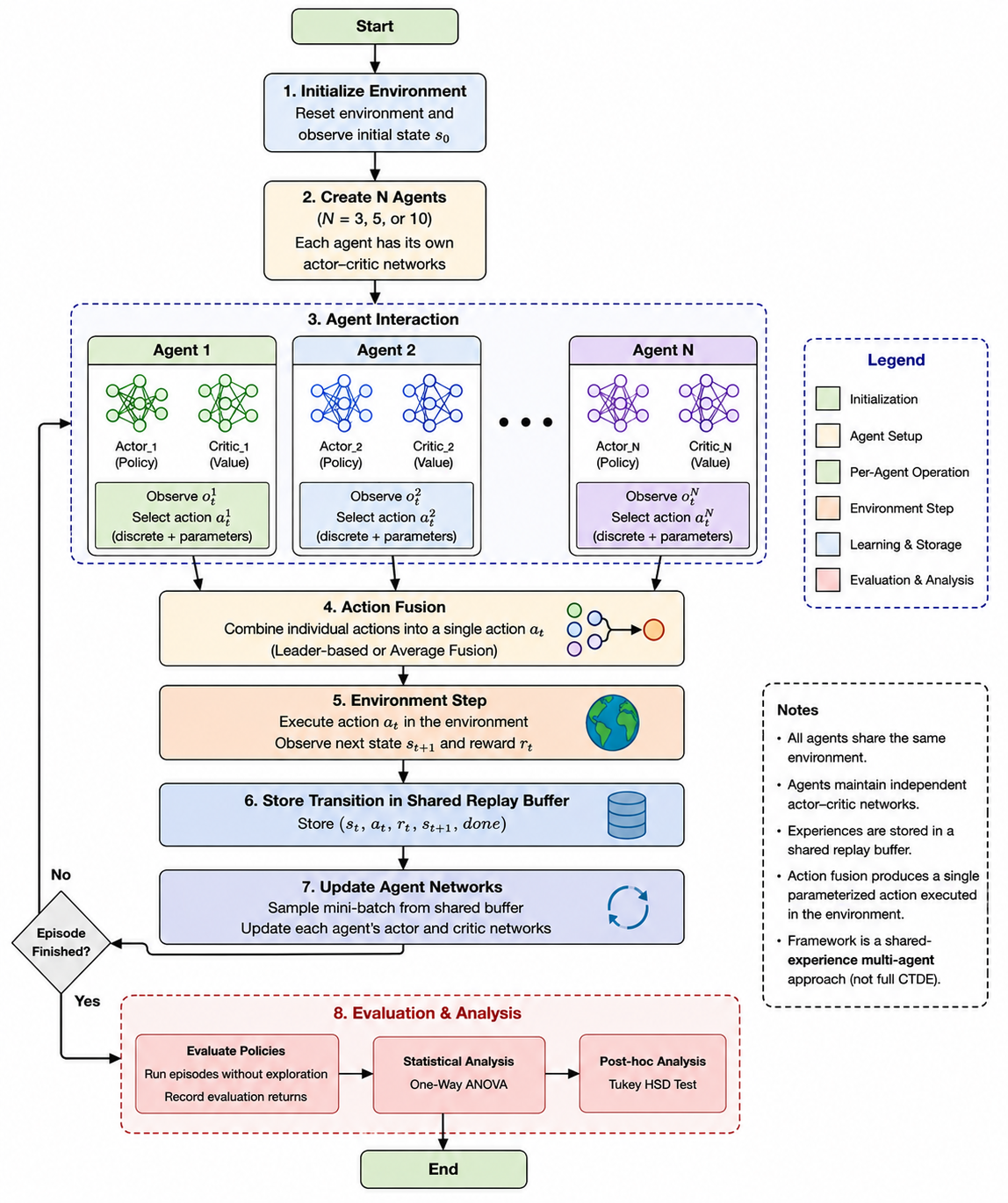}
\caption{Overview of the proposed shared-experience multi-agent actor--critic framework. Multiple actor--critic agents interact with a common environment, share experience through a replay buffer, and update their networks independently. The framework does not employ a centralized critic and therefore should be interpreted as a shared-experience multi-agent architecture rather than a full CTDE implementation.}
\label{fig:framework}
\end{figure}

Unlike centralized training and decentralized execution (CTDE) approaches commonly used in multi-agent reinforcement learning, the proposed framework does not employ a centralized critic operating on joint observations and joint actions. Instead, it adopts a lightweight shared-experience architecture in which agents learn through a common replay buffer while maintaining independent actor--critic networks. Therefore, the framework should be interpreted as a shared-experience multi-agent extension rather than a full CTDE implementation. Figure~\ref{fig:framework} illustrates the overall workflow of the proposed framework.

The training process begins by initialising the benchmark environment and creating the required number of agents. During each episode, the agents observe the current state and independently generate parameterised actions. These actions are fused into a single executable action, which is applied to the environment. The resulting transition is stored in the shared replay buffer and used to update the learning agents according to the underlying actor--critic algorithm. After training, the learned policies are evaluated using average evaluation return, training time, and statistical analysis.

\subsection{Agent Scalability}

To investigate the influence of the number of participating agents on learning performance, experiments were conducted using configurations consisting of three, five, and ten agents. All experiments used the same benchmark environments, hyperparameter settings, and training procedure, with only the number of agents varied. This allows the effect of increasing the number of agents on policy performance, computational efficiency, and scalability to be analysed independently

\section{Simulation Setup}

The experiments were conducted to evaluate the performance of actor–critic reinforcement learning algorithms operating in parameterised action environments. Three reinforcement learning algorithms were evaluated: Soft Actor-Critic (SAC), Greedy Actor-Critic (GAC), and Truncated Quantile Critics (TQC). In addition to their single-agent implementations, multi-agent extensions of these algorithms were also implemented, resulting in the algorithms MASAC, MAGAC, and MATQC.

The experiments were conducted using the benchmark environments Platform-v0 and Goal-v0 from the parameterised action reinforcement learning framework introduced by Bester et al. These environments require agents to select discrete actions together with associated continuous parameters, creating a hybrid action space that poses additional challenges for reinforcement learning algorithms. The baseline GAC, SAC and TQC implementations from our previous work were retained as reference algorithms for comparison with the proposed multi-agent implementations.

Hyperparameter optimisation was performed using Microsoft Neural Network Intelligence (NNI). To ensure reproducibility and fairness of comparison, all algorithms were implemented using the same codebase with modifications applied only to the actor–critic update mechanisms required for each algorithm. Each algorithm was evaluated across ten independent training runs in order to account for stochastic variations during training. The experiments were conducted for a total of 5,000 episodes per run. Performance was evaluated using the following metrics:

\begin{itemize}
\item Average evaluation return
\item Average training time
\item Statistical significance using One-Way ANOVA
\item Pairwise comparison using Tukey HSD
\end{itemize}

The evaluation return was calculated by averaging the final evaluation score between multiple runs, while training time was measured using the total duration required to complete each experiment.

\subsection{Experimental Configuration}

To ensure a fair comparison between the evaluated algorithms, all experiments were conducted using identical benchmark environments, training procedures and hyperparameter settings. The only variables modified during experimentation were the reinforcement learning algorithm and the number of participating agents.

The proposed multi-agent algorithms were evaluated using three different agent configurations consisting of three, five and ten agents. Each configuration was independently trained and evaluated on both the Platform-v0 and Goal-v0 benchmark environments.

To account for the stochastic nature of reinforcement learning, each experiment was repeated ten times using different random seeds and initial weight initialisations. The reported evaluation returns and training times therefore represent the average performance obtained across ten independent experimental runs. Algorithm performance was assessed using two primary evaluation metrics:

\begin{itemize}
\item Average evaluation return
\item Average training time
\end{itemize}

The average evaluation return was used to assess the quality of the learned policy, while the average training time was used to evaluate the computational efficiency of each algorithm. To determine whether the observed differences between algorithms were statistically significant, one-way Analysis of Variance (ANOVA) was performed for each benchmark environment. Where statistically significant differences were identified, Tukey Honest Significant Difference (HSD) post-hoc testing was conducted to identify the specific algorithm pairs responsible for the observed differences.

\section{Results}

This section presents the experimental results obtained from the evaluation of the proposed multi-agent actor--critic algorithms. The algorithms were evaluated using both the Goal-v0 and Platform-v0 benchmark environments. Performance was assessed using average evaluation return and average training time across ten independent experimental runs for each configuration. To investigate the influence of the number of participating agents, experiments were conducted using three, five and ten agents. 

\subsection{Performance Comparison}

Tables~\ref{tab:goal_summary} and~\ref{tab:platform_summary} compare the performance of the original single-agent algorithms (GAC, SAC and TQC) with the proposed multi-agent implementations evaluated using three-, five-, and ten-agent configurations. Performance is reported using the average evaluation return and the average training time across ten independent experimental runs. The results demonstrate that extending the actor--critic algorithms to a shared-experience multi-agent framework does not uniformly improve performance across all algorithms. Instead, the impact of increasing the number of participating agents depends on the underlying reinforcement learning algorithm and the benchmark environment.

\begin{table}[ht]
\centering
\caption{Summary of experimental results for the Goal environment.}
\label{tab:goal_summary}
\begin{tabular}{lccc}
\hline
Algorithm & Configuration & Avg Return & Avg Time (hh:mm:ss) \\
\hline
GAC   & Baseline  & -8.551 & 00:24:05 \\
MAGAC & 3 Agents & -7.914 & 00:37:49 \\
MAGAC & 5 Agents & -7.885 & 01:30:11 \\
MAGAC & 10 Agents & -7.893 & 02:55:18 \\
\hline
SAC   & Baseline  & -8.200 & 00:26:37 \\
MASAC & 3 Agents & -8.239 & 00:17:47 \\
MASAC & 5 Agents & -8.268 & 00:24:44 \\
MASAC & 10 Agents & -8.188 & 00:23:20 \\
\hline
TQC   & Baseline  & -8.155 & 00:27:25 \\
MATQC & 3 Agents & -8.225 & 00:12:49 \\
MATQC & 5 Agents & -8.109 & 00:13:27 \\
MATQC & 10 Agents & -8.120 & 00:18:18 \\
\hline
\end{tabular}
\end{table}

Table~\ref{tab:goal_summary} shows that the proposed multi-agent extension provides the greatest benefit for the Greedy Actor-Critic algorithm. The baseline GAC achieved an average evaluation return of -8.551, while MAGAC improved this value to -7.914 using three agents and further to -7.885 using five agents. Increasing the number of agents to ten produced virtually no additional improvement, indicating diminishing returns beyond five agents.

\begin{table}[ht]
\centering
\caption{Summary of experimental results for the Platform environment.}
\label{tab:platform_summary}
\begin{tabular}{lccc}
\hline
Algorithm & Configuration & Avg Return & Avg Time (hh:mm:ss) \\
\hline
GAC   & Baseline  & 0.170 & 00:41:23 \\
MAGAC & 3 Agents & 0.183 & 01:31:29 \\
MAGAC & 5 Agents  & 0.185 & 02:53:32 \\
MAGAC & 10 Agents & 0.185 & 06:00:01 \\
\hline
SAC   & Baseline  & 0.143 & 00:56:16 \\
MASAC & 3 Agents & 0.145 & 00:42:18 \\
MASAC & 5 Agents & 0.146 & 00:50:50 \\
MASAC & 10 Agents & 0.146 & 00:52:30 \\
\hline
TQC   & Baseline  & 0.145 & 00:51:40 \\
MATQC & 3 Agents & 0.148 & 00:27:15 \\
MATQC & 5 Agents & 0.144 & 00:29:47 \\
MATQC & 10 Agents & 0.146 & 00:32:14 \\
\hline
\end{tabular}
\end{table}

Table~\ref{tab:platform_summary} demonstrates similar behaviour within the Platform environment. The transition from the baseline GAC algorithm to MAGAC resulted in a clear improvement in average evaluation return, increasing from 0.170 to 0.183 using three agents and 0.185 using both five and ten agents. However, no measurable performance improvement was observed beyond five agents despite a considerable increase in computational cost.

The multi-agent extensions of SAC and TQC again exhibited relatively modest performance changes compared to their corresponding baseline implementations. MASAC maintained consistent evaluation returns across all evaluated configurations, while MATQC achieved only minor improvements regardless of the number of participating agents. Training time increased substantially for MAGAC as the number of agents increased, reaching approximately six hours for the ten-agent configuration. In comparison, MASAC and MATQC maintained significantly lower computational requirements across all evaluated configurations. These findings indicate that increasing the number of participating agents beyond five provides limited additional performance improvements while substantially increasing computational cost. The Goal environment demonstrates that MAGAC achieved the highest evaluation performance among the evaluated configurations, particularly when increasing the number of participating agents. MASAC and MATQC exhibited comparatively smaller performance improvements as the number of agents increased.

A similar trend was observed within the Platform environment. Although additional agents occasionally improved evaluation return, the improvements were generally modest and did not always justify the increased computational cost associated with larger multi-agent configurations.

\subsection{Training Time Analysis}

\begin{figure}[ht]
\centering
\includegraphics[width=\linewidth]{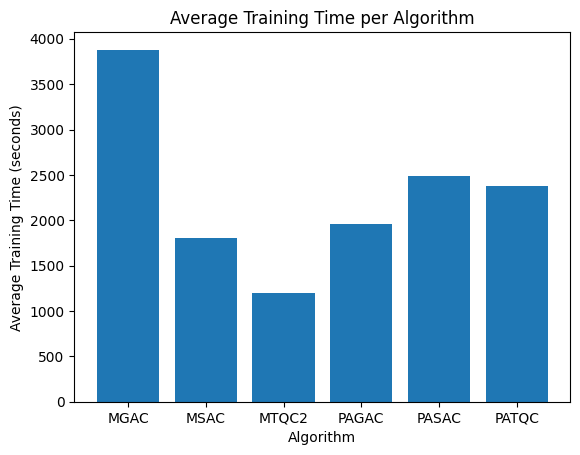}
\caption{Average training time for each evaluated algorithm.}
\label{fig:training_time}
\end{figure}

Figure~\ref{fig:training_time} illustrates the average training time required by each evaluated algorithm.

Increasing the number of participating agents increased the computational cost of training for all evaluated algorithms. This increase was particularly evident for MAGAC, where training time increased substantially as the number of participating agents grew. Although larger multi-agent configurations occasionally produced improved evaluation returns, the increase in computational cost was not proportional to the observed performance improvements. These results suggest diminishing computational efficiency as the number of participating agents increases.

\subsection{Goal Environment Results}

The evaluation results obtained from the Goal environment experiments reveal differences in the behaviour of the evaluated multi-agent algorithms. While all algorithms successfully learned policies capable of achieving stable evaluation returns, variations were observed in the resulting performance metrics. MAGAC demonstrated distinct performance characteristics compared to MASAC and MATQC. This behaviour suggests that the greedy policy improvement mechanism used in the GAC framework influences the learning dynamics of the algorithm.

To formally evaluate whether the observed performance differences were statistically significant, a one-way ANOVA analysis was conducted using the evaluation returns obtained from multiple experimental runs.

\subsection{ANOVA Analysis for the Goal Environment}

Table \ref{tab:goal_anova} presents the ANOVA results for the multi-agent algorithms in the Goal environment.

\begin{table}[h]
\centering
\caption{One-Way ANOVA results for multi-agent algorithms in the Goal environment}
\label{tab:goal_anova}
\resizebox{\columnwidth}{!}{%
\begin{tabular}{lccccc}
\hline
Source & DF & Sum of Squares & Mean Square & F Statistic & p-value \\
\hline
Between Groups & 2 & 0.6748 & 0.3374 & 32.97 & $5.659\times10^{-8}$ \\
Within Groups & 27 & 0.2763 & 0.01023 & & \\
Total & 29 & 0.9511 & & & \\
\hline
\end{tabular}%
}
\end{table}

The ANOVA analysis produced an F-statistic of $F(2,27)=32.97$ with a corresponding p-value of $5.659 \times 10^{-8}$. Since the p-value is significantly smaller than the significance threshold ($\alpha = 0.05$), the null hypothesis was rejected. This indicates that statistically significant differences exist between the evaluated algorithms. The effect size measured using $\eta^2$ was $0.71$, indicating that approximately 70.9\% of the variance in evaluation performance can be attributed to differences between the algorithms.

\subsection{Tukey Post-Hoc Analysis for the Goal Environment}

Although the ANOVA analysis confirms the presence of statistically significant differences, it does not identify which algorithms differ from one another. Therefore, a Tukey Honest Significant Difference (HSD) post-hoc test was performed.

\begin{table}[h]
\centering
\caption{Tukey HSD post-hoc test results for the Goal environment}
\label{tab:goal_tukey}
\resizebox{\columnwidth}{!}{%
\begin{tabular}{lcccccc}
\hline
 Pair & Difference & SE & Q & Lower CI & Upper CI & p-value \\
\hline
MAGAC -- MASAC & 0.3248 & 0.03199 & 10.15 & 0.2126 & 0.4369 & $2.99\times10^{-7}$ \\
MAGAC -- MATQC & 0.3111 & 0.03199 & 9.72 & 0.1989 & 0.4233 & $6.41\times10^{-7}$ \\
MASAC -- MATQC & 0.0137 & 0.03199 & 0.43 & -0.0985 & 0.1258 & 0.951 \\
\hline
\end{tabular}
}
\end{table}

The Tukey HSD results indicate that MAGAC differs significantly from both MASAC and MATQC. In contrast, the comparison between MASAC and MATQC was not statistically significant, indicating similar performance between these algorithms. These results suggest that the greedy policy update mechanism used in the GAC framework introduces distinct learning behaviour when compared with entropy-based and distributional reinforcement learning approaches.

\subsection{Platform Environment Results}

The experimental results obtained from the Platform environment experiments demonstrate similar patterns to those observed in the Goal environment. While all algorithms were able to learn stable policies, differences in evaluation performance were observed across the evaluated multi-agent algorithms. MAGAC again demonstrated distinct performance characteristics compared to MASAC and MATQC. To determine whether these differences were statistically significant, an ANOVA analysis was conducted.

\subsection{ANOVA Analysis for the Platform Environment}

Table \ref{tab:platform_anova} presents the ANOVA results for the Platform environment.

\begin{table}[h]
\centering
\caption{One-Way ANOVA results for multi-agent algorithms in the Platform environment}
\label{tab:platform_anova}
\resizebox{\columnwidth}{!}{%
\begin{tabular}{lccccc}
\hline
Source & DF & Sum of Squares & Mean Square & F Statistic & p-value \\
\hline
Between Groups & 2 & 0.009139 & 0.004569 & 100.80 & $2.99\times10^{-13}$ \\
Within Groups & 27 & 0.001224 & 0.00004533 & & \\
Total & 29 & 0.01036 & & & \\
\hline
\end{tabular}
}
\end{table}

The ANOVA analysis produced an F-statistic of $F(2,27)=100.80$ with a p-value of $2.99 \times 10^{-13}$. This indicates extremely strong statistical evidence that the evaluated algorithms exhibit different performance characteristics. The effect size $\eta^2 = 0.88$ indicates that approximately 88.2\% of the variance in evaluation performance can be explained by differences between the algorithms.

\subsection{Tukey Post-Hoc Analysis for the Platform Environment}

\begin{table}[h]
\centering
\caption{Tukey HSD post-hoc test results for the Platform environment}
\label{tab:platform_tukey}
\resizebox{\columnwidth}{!}{%
\begin{tabular}{lcccccc}
\hline
Pair & Difference & SE & Q & Lower CI & Upper CI & p-value \\
\hline
MAGAC -- MASAC & 0.0381 & 0.002129 & 17.90 & 0.03064 & 0.04557 & $4.64\times10^{-12}$ \\
MAGAC -- MATQC & 0.03584 & 0.002129 & 16.83 & 0.02838 & 0.04331 & $1.14\times10^{-11}$ \\
MASAC -- MATQC & 0.00226 & 0.002129 & 1.06 & -0.00521 & 0.00973 & 0.736 \\
\hline
\end{tabular}
}
\end{table}

The Tukey HSD test confirms that MAGAC differs significantly from both MASAC and MATQC. The comparison between MASAC and MATQC remains statistically insignificant, indicating that these algorithms demonstrate similar performance behaviour. Overall, the results obtained from both environments show consistent patterns in algorithm behaviour. MAGAC consistently demonstrates statistically significant differences compared to MASAC and MATQC, while MASAC and MATQC produce comparable results.

\section{Discussion}

The objective of this study was to investigate whether extending parameterised action actor--critic reinforcement learning algorithms to a shared-experience multi-agent framework improves learning performance within parameterised action environments. The experimental results demonstrate that the proposed multi-agent implementations successfully learn effective policies in both the Goal-v0 and Platform-v0 benchmark environments. However, the magnitude of improvement varies across the evaluated algorithms. Among the evaluated algorithms, MAGAC consistently demonstrated the greatest improvement over its corresponding single-agent implementation. In both benchmark environments, the transition from GAC to MAGAC produced higher evaluation returns, while increasing the number of agents from three to five yielded only marginal additional improvements. Increasing the number of participating agents to ten provided virtually no measurable performance benefit, indicating diminishing returns beyond moderate-sized multi-agent configurations.

In contrast, MASAC and MATQC exhibited relatively stable performance across all evaluated configurations. Both algorithms produced evaluation returns that remained close to those of their corresponding single-agent implementations, suggesting that increasing the number of participating agents has only a limited influence on learning performance for these approaches. This trend was consistently observed across both benchmark environments.

The scalability experiments also revealed a clear trade-off between policy performance and computational cost. While MAGAC achieved the highest evaluation returns, its training time increased substantially as the number of agents increased. Conversely, MASAC and MATQC maintained comparatively low computational costs while producing relatively small changes in evaluation performance. These findings demonstrate that increasing the number of participating agents alone does not necessarily improve learning performance and that algorithm selection has a greater influence than simply increasing agent count. The One-Way ANOVA and Tukey HSD analyses support these observations by confirming statistically significant performance differences between the evaluated multi-agent algorithms. In particular, MAGAC differed significantly from MASAC and MATQC in both benchmark environments, while MASAC and MATQC frequently exhibited statistically similar behaviour.

This observation is consistent with the broader multi-agent reinforcement learning literature, where increasing the number of agents often improves exploration and coordination opportunities but also introduces additional communication, optimisation, and computational overhead. Similar scalability challenges have been reported for approaches such as QMIX \cite{rashid2018qmix} and MAPPO \cite{yu2022mappo}, where coordination quality and computational complexity become increasingly important as the number of agents grows.

This work extends the comparative analysis presented in ~\cite{bapoo2025icit}, which evaluated the corresponding single-agent parameterised action reinforcement learning algorithms. Unlike centralized training and decentralized execution (CTDE) approaches, the proposed framework employs independent actor--critic agents that learn through a shared replay buffer without a centralized critic. Future work will investigate centralized critics for parameterised action spaces, parameter sharing between agents, alternative action fusion strategies, and evaluation within more complex cooperative multi-agent benchmark environments.

\section{Conclusion}

This paper presented a comparative study of three multi-agent parameterised action actor--critic reinforcement learning algorithms, namely Multi-Agent Greedy Actor-Critic (MAGAC), Multi-Agent Soft Actor-Critic (MASAC), and Multi-Agent Truncated Quantile Critics (MATQC). Building upon previous work on the corresponding single-agent algorithms, this study investigated the effect of extending these approaches to a shared-experience multi-agent framework within parameterised action reinforcement learning environments. The proposed algorithms were evaluated using the Goal-v0 and Platform-v0 benchmark environments and compared with the original GAC, SAC and TQC implementations. Scalability was investigated using three-, five-, and ten-agent configurations, while performance differences were assessed using average evaluation return, training time, One-Way ANOVA, and Tukey HSD statistical analysis.

The experimental results demonstrate that the proposed multi-agent framework benefits the Greedy Actor-Critic algorithm more than the Soft Actor-Critic and Truncated Quantile Critics algorithms. While MAGAC consistently achieved the highest evaluation performance, increasing the number of participating agents beyond five produced only marginal improvements while substantially increasing computational cost. These findings suggest that moderate multi-agent configurations provide a favourable balance between learning performance and computational efficiency.

Although the proposed framework is not a full centralized training and decentralized execution (CTDE) implementation, it demonstrates the potential of shared-experience multi-agent learning for parameterised action reinforcement learning. Future work will investigate centralized critics, parameter sharing between agents, and more complex cooperative multi-agent environments to further improve scalability and learning performance. In particular, future work will explore value-decomposition approaches such as QMIX \cite{rashid2018qmix} together with policy-gradient methods including MAPPO \cite{yu2022mappo} to determine whether stronger coordination mechanisms further improve learning within parameterised action spaces.

\bibliographystyle{IEEEtran}
\bibliography{references}

\end{document}